%% file: main.tex
\documentclass[10pt,twocolumn,letterpaper]{article}

\usepackage[pagenumbers]{cvpr} 

\input{preamble}
\definecolor{cvprblue}{rgb}{0.21,0.49,0.74}
\usepackage[pagebackref,breaklinks,colorlinks,allcolors=cvprblue]{hyperref}
\usepackage{pifont}
\def\paperID{19690} 
\def\confName{CVPR}
\def\confYear{2026}

\title{Determined by User Needs: A Salient Object Detection Rationale Beyond Conventional Visual Stimuli}
\author{Chenglizhao Chen\\
China University of Petroleum (East China)\\
\and
Shujian Zhang\\
China University of Petroleum (East China)\\
\and
Luming Li$^{*}$\\
China University of Petroleum (East China)\\
Wenfeng Song\\
Beijing Information Science and Technology University\\
\and
Shuai Li\\
Beihang University \\
}

\begin{document}
\maketitle
\input{sec/0_abstract}    
\input{sec/1_intro}
\input{sec/2_Related}
\input{sec/3_Method}
\input{sec/4_Experiment}
\input{sec/5_Concluison}
{
    \small
    \bibliographystyle{ieeenat_fullname}
    \bibliography{main}
}


\end{document}

%% file: sec/0_abstract.tex
\begin{abstract}
Existing \textbf{s}alient \textbf{o}bject \textbf{d}etection (SOD) methods adopt a \textbf{passive} visual stimulus-based rationale--objects with the strongest visual stimuli are perceived as the user's primary focus (i.e., salient objects).
They ignore the decisive role of users' \textbf{proactive needs} in segmenting salient objects--if a user has a need before seeing an image, the user's salient objects align with their needs, e.g., if a user's need is ``white apple'', when this user sees an image, the user's primary focus is on the ``white apple'' or ``the most white apple-like'' objects in the image.
Such an oversight not only \textbf{fails to satisfy users}, but also \textbf{limits the development of downstream tasks}.
For instance, in salient object ranking tasks, focusing solely on visual stimuli-based salient objects is insufficient for conducting an analysis of fine-grained relationships between  users' viewing order (usually determined by user's needs) and scenes, which may result in wrong ranking results.
Clearly, it is essential to detect salient objects based on user needs.
Thus, we advocate a \textbf{User} \textbf{S}alient \textbf{O}bject \textbf{D}etection (UserSOD) task, which focuses on \textbf{detecting salient objects align with users' proactive needs when user have needs}.
The main challenge for this new task is the lack of datasets for model training and testing.
We propose a labor-efficient and cost-effective sample digging protocol to construct a reliable dataset.
Meanwhile, to the best of our knowledge, there is no existing model for this emerging task.
To fill this gap, we develop a model named UserSal.
Further, to enhance UserSal's ability to handle users' fine-grained object-related needs (e.g., color, shape, and more), we propose a component incorporating fine-grained information and the corresponding loss.
Experiments demonstrate that our UserSOD task, sample protocol, dataset, model, component, and loss can advance the field.
\end{abstract}

%% file: sec/1_intro.tex
\section{Introduction}
\label{sec:intro}
Salient Object Detection (SOD) is a computer vision task that simulates the user visual system's selective attention mechanism, aiming to segment the most noticeable objects (\emph{i.e.}, salient objects) in various scenes.
It plays a significant role in downstream tasks such as image retrieval~\cite{10103906}, activity recognition~\cite{10003254}, and salient object ranking~\cite{qiao2024hypersor}.
\begin{figure}[!t]
	\centering
	\includegraphics[width=1\linewidth]{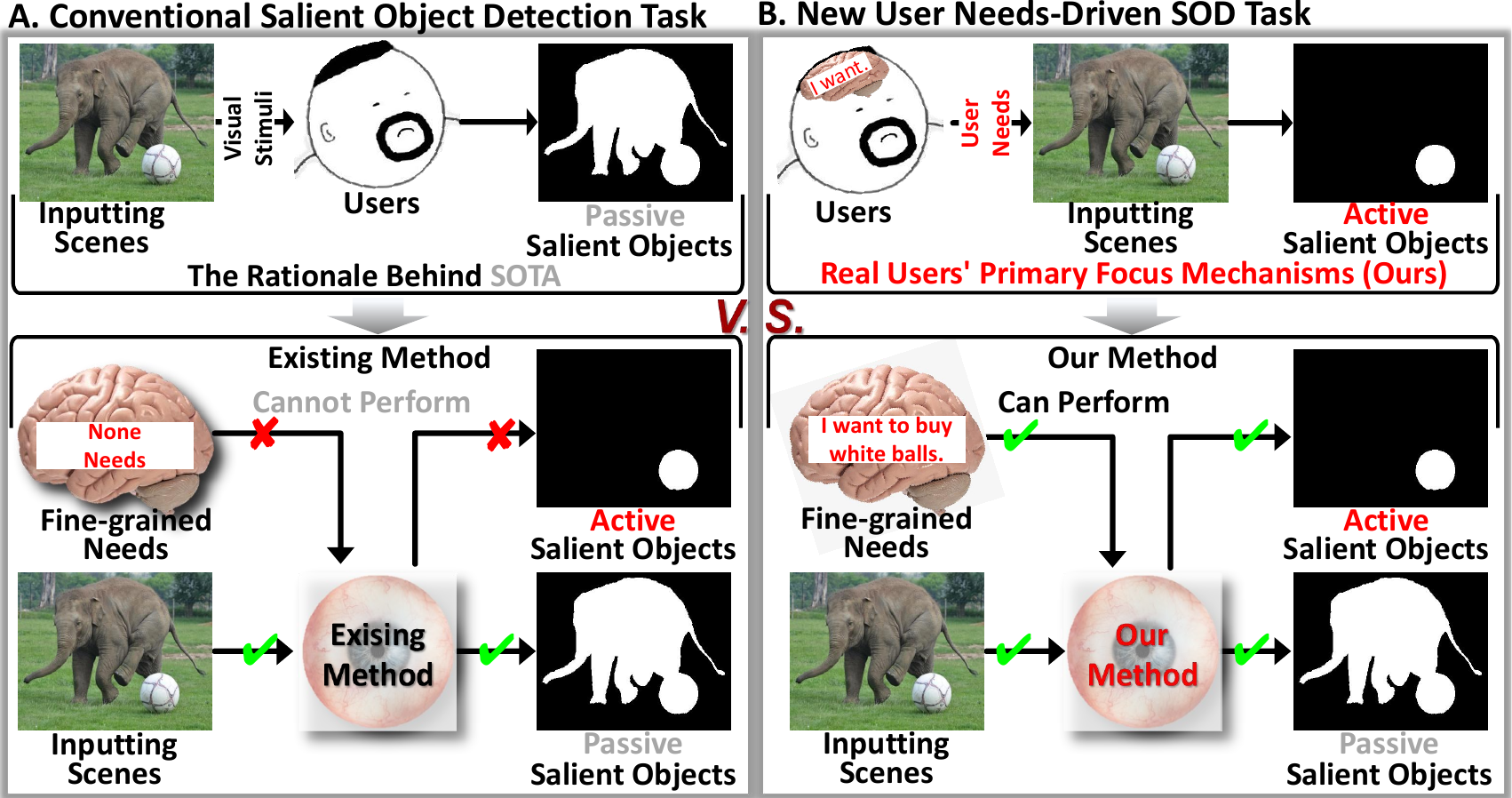}
	\caption{Motivation demonstration of our new task. Current SOTA methods (A) always segment salient object based on passive visual stimulus-based rationales, limiting the performance of downstream tasks (\emph{e.g.}, salient object ranking, see details in the Fig.~\ref{fig:subtask}). Therefore, we introduce a novel task, UserSOD (B), which detects salient object based on user need, as it accounts for the crucial role of user active need, thereby aligning more closely with real users' primary focus mechanisms.}
	\label{fig:motivation}
\end{figure}

Existing SOD methods~\cite{Han-TYCB18-CTMF,Pang-CVPR20-Mul,MobileSal} segment salient objects based on visual stimulus 
rationales, regarding those that are most likely to attract users' visual attention, \emph{e.g.}, objects with the brightest color, objects with the largest size, as users' salient objects (Fig.~\ref{fig:motivation}(A)) in scenes.
While this passive detection rationale aligns with fundamental human physiological mechanisms, it fails to account for the crucial role of users' active need in determining salient objects (Fig.~\ref{fig:motivation}(B)), \emph{i.e.},
when actively searching for objects, users preferentially focusing on objects that match their current needs.
This user active needs-driven SOD rationale becomes particularly prominent in downstream tasks.
\begin{figure}[!t]
	\centering
	\includegraphics[width=1\linewidth]{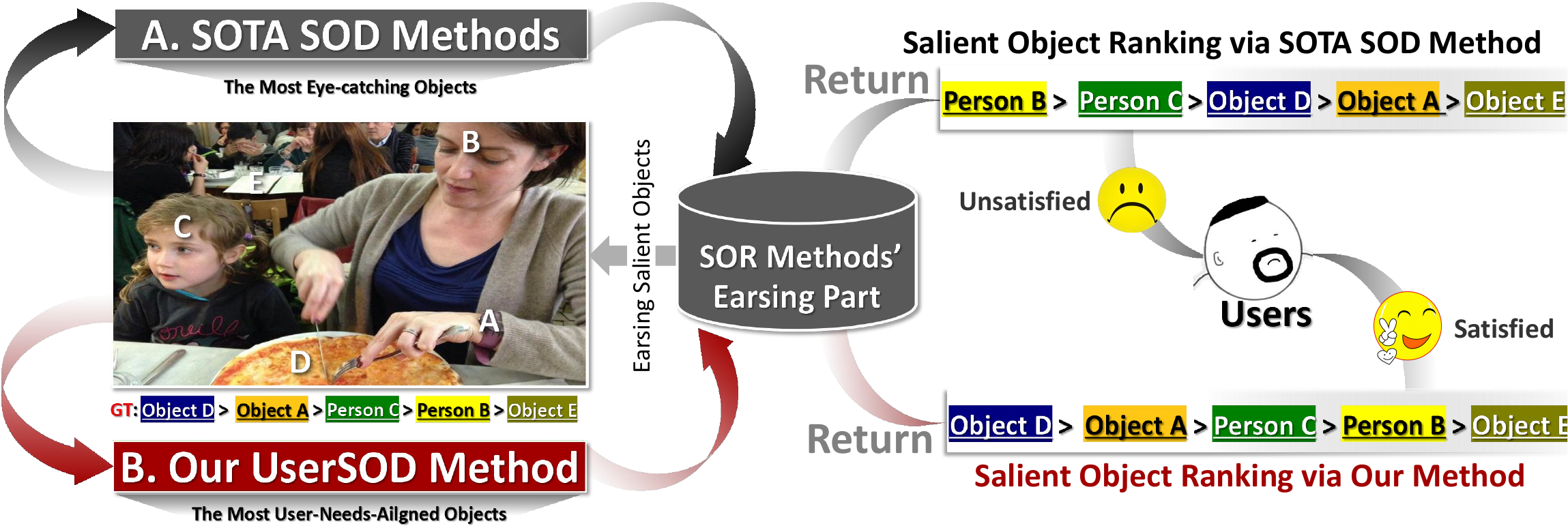}
	\caption{
		Application of salient object detection method to salient object ranking (SOR).
       (A) The existing SOTA SOD methods ignores user active need, leading to constant rank objects' saliency that fail to satisfy the user needs and deviate from the ground truth (GT, Bottom left).
       In contrast, our approach (B) can return more fine-grained rank results, better meeting user needs.
	}
	\label{fig:subtask}
\end{figure}

For instance, as shown in Fig.~\ref{fig:subtask}, the \textbf{s}alient \textbf{o}bject \textbf{r}anking (SOR) task is defined as the ranking of objects' saliency to the users in a scene.
The core challenge lies in analyzing the fine-grained relationship between users' viewing behaviors and scenes, specifically, users' viewing order of objects in a scene.
However, users' viewing order is not constant, even for the same user and scene.
Existing SOTA methods (Fig.~\ref{fig:subtask}(A)) employ passive visual stimuli-based SOD methods to simulate the users' viewing orders, \emph{e.g.},
they iteratively cycle through the most visually salient objects in the scene---detecting salient objects---erasing---detecting, subsequently returning ranking results based on fixed stimuli.
However, these SOTA methods ignore the variant nature of users' viewing order.
This coarse-grained analysis may produce wrong ranking results that significantly deviate from the ground truth and fail to meet users' needs.
Therefore, we propose a new task, namely \textbf{User} \textbf{S}alency \textbf{O}bject \textbf{D}etection (UserSOD), which focuses on detecting salient objects based on user active needs\footnote{User needs-driven salient objects are the objects most aligned with user needs in an image. For example, if a user wants to find a red car and the image contains a red car, it is the salient object; if not, the salient object would be the most red car-like object.}.
By applying our UserSOD, we can capture the fine-grained relationship users' viewing order and scenes,
which substantially improves the performance of SOR methods (Fig.~\ref{fig:subtask}(B)).

As a new task\footnote{The clear distinction between our task and similar multi-modal segmentation tasks such as referring image segmentation is that, in our work, the text represents user needs, while in other tasks it describes the image in detail.}, UserSOD's first challenge is the lack of training and testing samples.
Specifically, large-scale sample pairs with images corresponding to user need commands are difficult to obtain due to collection challenges\footnote{Users' needs vary across different times and locations, and the salient objects corresponding to these needs also differ. Manual annotation and collection would evidently require enormous labor and financial costs}.
To address this, we propose a sample dig protocol that reduces the dependency on manually collecting samples.
This protocol enables semi-automatic mining of high-quality user need commands that are aligned with images for model training and testing.
By leveraging these samples, the community can develop user needs-driven SOD models that achieve superior performance.

UserSOD's second challenge is the lack of a model for detecting salient objects under user with needs.
As a pioneering effort, we propose a novel unified framework---Usersal---that explicitly prioritizes user needs in SOD by leveraging well-known \textbf{m}ulti-modal \textbf{p}rompt \textbf{l}earning (MPL) tools.
Based on this model, Usersal can perform both UserSOD and conventional SOD tasks.
Further, to address the limitation of MPL that cannot make models perceive fine-grained cues, we propose a component and corresponding loss to incorporate fine-grained cues from user need commands,
which can better satisfy users.
In summary, our main contributions are:
\begin{itemize}
\item We introduce UserSOD, a novel task aimed at detecting salient objects based on user needs, which can benefit downstream tasks.
\item We propose an innovative sample dig protocol for the UserSOD task, effectively addressing the challenge of sample scarcity.
     Additionally, we construct a large-scale dataset---UserSOD---containing numerous images, user need commands, and corresponding saliency maps (\emph{i.e.}, ground truth).
\item For this new task, we develop a unified model that performs both UserSOD and conventional SOD tasks.
\item We propose a component and corresponding loss for incorporating fine-grained cues from user need commands to better satisfy users.
\item We will publicly release the dataset, model, and experimental results to advance research in this field.
\end{itemize}

%% file: sec/2_Related.tex
\section{Related Work}
\label{sec:RW}
\subsection{Salient Object Detection (SOD)}
SOD aims to predict the most noticeable object of users within a image.
State-of-the-art (SOTA) SOD research~\cite{TSD,MENet,EDN} primarily focuses on enhancing models' feature representation capabilities to improve the performance of SOD methods.
For example:
Pang \emph{et al.}~\cite{Pang-CVPR20-Mul} have leveraged the powerful feature representation of CNNs to capture both low-level and high-level information.
Liu \emph{et al.}~\cite{Liu-arxiv21-VST} have employed a Transformer-based encoder-decoder architecture to model long-range global dependencies, resulting in more discriminative salient object features.
Chen \emph{et al.}~\cite{diffCOD} have utilized a latent diffusion model to extract high-level semantic features.
Despite significant progress in SOD, existing methods adhere to a \textbf{passive} rationale---where salient objects are determined by visual stimulus,
neglecting the critical influence of user needs in object saliency judgment.
This oversight fundamentally limits the applicability and development of SOD-based downstream tasks.
To address, we propose UserSOD, an \textbf{active} rationale that aligns with user attention mechanisms—--salient objects as those most semantically relevant to user needs.
\subsection{Multi-modal Prompt Learning (MPL)}
By appending extra token prompts derived from samples of other modalities to pre-frozen existing models, MPL~\cite{CLIP} achieves unified representation learning for visual, linguistic, and other multi-modal data.
For instance, Radford \emph{et al.}~\cite{CLIP} have introduced contrastive learning into MPL, aligning image-text embeddings into a shared feature space to enhance the model’s representation for multi-modal data.
BLIP~\cite{BLIP} automatically cleanses noisy acquired data to obtain high-quality image-text pairs, enabling improved multi-modal feature representation with minimal data.
FLIP~\cite{FLIP} combines masked image modeling with contrastive learning, not only enhancing multi-modal representation but also drastically reducing computational costs.
While we adapt MPL to transform existing SOD models into ones can perform both passive SOD and UserSOD tasks, current MPL frameworks lack the ability to make models perceive fine-grained user needs.
This work addresses this critical limitation by proposing appearance similar adder, which we will detail in Sec.~\ref{sec:model}.
\begin{figure}[!t]
	\centering
	\includegraphics[width=\linewidth]{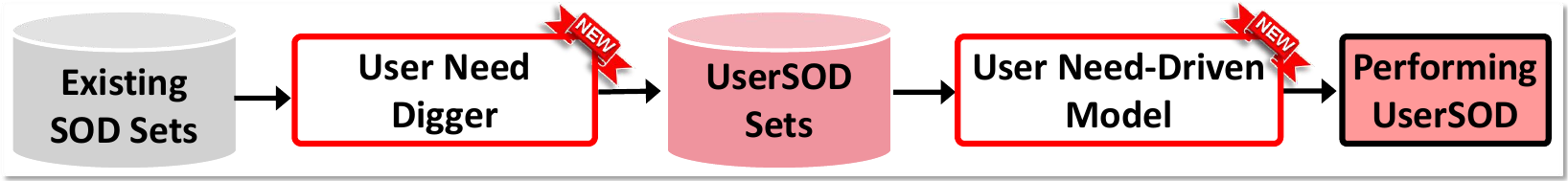}
	\caption{The pipeline of our method. Our consists of two components, \emph{i.e.}, user need digger (UND) and user need-driven salient object detection model (Uersal$^+$).
UND semi-automatically digs user need commands based on existing samples, and Uersal$^+$ can leverage fine-grained clues from user need commands to detect salient objects.}
	\label{fig:pipeline}
\end{figure}

%% file: sec/3_Method.tex
\section{Method}
Existing salient object detection (SOD) methods adopt a \textbf{passive} rationale, limiting developments of downstream tasks.
Thus, we propose the UserSOD task.
The pipeline are shown in Fig.~\ref{fig:pipeline}; the core idea is to dig and leverage user needs.
To address the sample shortage problem, we propose a User Need Digger (Sec.~\ref{sec:URD}), which semi-automatically digs user need for samples contained in conventional SOD dataset,
thereby it is  a money-consuming and labor-intensive.
Based on these samples, we train a model capable of utilizing user fine-grained needs (Usersal$^+$, Sec.~\ref{sec:Usersal}) to segment salient objects.
\subsection{User Need Digger: A Semi-automatical Tool}
\label{sec:URD}
\begin{figure}[!t]
	\centering
	\includegraphics[width=1\linewidth]{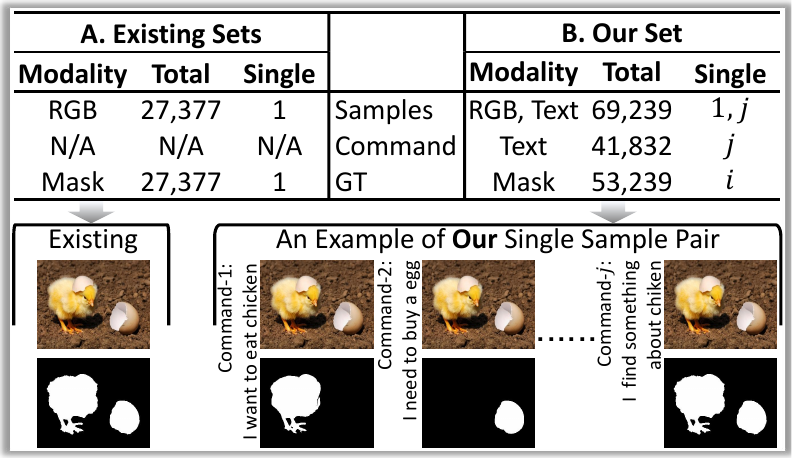}
	\caption{Existing Sets \emph{v.s.} Our UserSOD set. Compared to existing set, Users contains samples which are image, corresponding user need commands, and corresponding \textbf{g}round \textbf{t}ruths (GT).
Where, \emph{i}$\subset$(1, $+ \infty$) and \emph{j}$\subset$(1, $+ \infty$) denote the number of mask and user need command in single samples, respectively.
	}
	\label{fig:sets}
\end{figure}
One of the reasons why existing methods fail to detect salient objects driven by user needs:
existing sets (Fig.~\ref{fig:sets}(A)) don't have samples to train models---existing samples lack corresponding user need command and \textbf{g}round \textbf{t}ruths (GT).
However, collecting such samples in manual way is money-consuming and labor-intensive.
Thus, we should deal with an extremely difficult problem---\textbf{how to add a large number of correct user need command and corresponding GT for existing samples in a cost-effective way}?
We propose a semi-automatical digging scheme---\textbf{U}ser \textbf{N}eed \textbf{D}igger (UND), which leverages the anthropomorphic capabilities of SOTA foundational models~\cite{2025Human}.

\begin{figure}[!t]
	\centering
	\includegraphics[width=\linewidth]{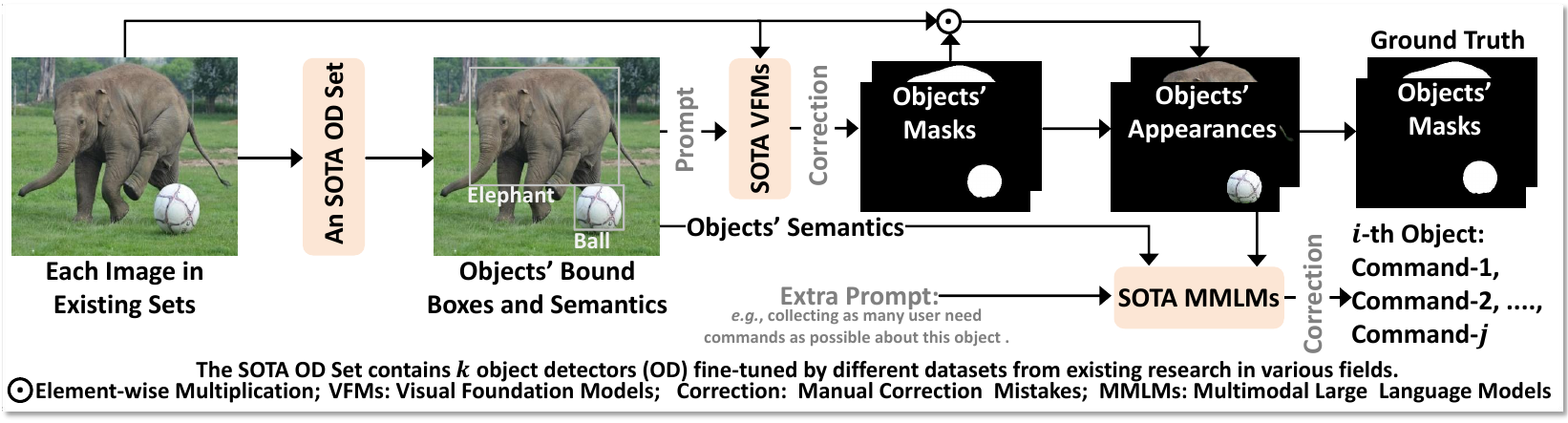}
	\caption{The proposed User Need Digger.}
	\label{fig:UTD}
\end{figure}

UND (Fig.~\ref{fig:UTD}) semi-automatically dig user need commands for existing samples based on three phase:
1)~digging latent target---digging use-needed latent target in existing sample; 2)~user needs inference--inferencing users' needs based on obtained latent targets; 3)~sample refinement--correct manually.
UND first needs to address how to ``find'' latent user need targets in existing samples.
Our solution is to leverage a set of \textbf{o}bject \textbf{d}etectorss\footnote{Object detectors are widely used in research across different fields. Most studies fine-tune them with datasets from their respective fields or collected independently, where the annotated objects can represent the objects needed by users in different fields. Otherwise, there is no need to annotate these objects.} (OD$^1$, OD$^2$,..., OD$^k$), \emph{e.g.}, YOLOv8~\cite{yolov8} fine-tuned by different datasets from existing research in various fields (\textit{k}, \emph{k}$\in$ (1, $+ \infty$)), including animal-related, agricultural, autonomous driving, and more field.
Thus, give an image ($\mathcal{\textbf{I}}$$\in$$\mathbb{R}^{H \times W}$) of existing samples, UND first performs Phase-1) to locate each object's \textbf{b}ounding \textbf{b}ox ($\textbf{\rm BB}_{i}$, \emph{i}$\in$(0, $+ \infty$)) and semantic ($\textbf{\rm O}_i$), which is achieved by OD$^1$-OD$^k$.
Based on $\textbf{\rm BB}_{i}$ and $\textbf{\rm O}_i$, UND~infers \textbf{u}ser \textbf{n}eed \textbf{c}ommand (UNC$_j$, \emph{j}$\in$(1, $+ \infty$)) and the \textbf{m}ask (M$_i$) of each latent target.
To get M$_i$, UND feeds $\mathcal{\textbf{I}}$ to existing \textbf{v}isual \textbf{f}oundational \textbf{m}odel (VFM($\cdot$), \emph{e.g.}, SAMv2~\cite{SAM}), and take $\textbf{\rm BB}_{i}$ as prompt for them.
Then, the M$_i$ are corrected manually and element-wisely multiplied by $\mathcal{\textbf{I}}$ to get each latent object's appearance.
Finally, UNC are corrected \textbf{man}ually (Man($\cdot$)) after feeding
each latent object's appearance, $\textbf{\rm O}_i$, and extra \textbf{p}ro\textbf{m}p\textbf{t} (Pmt) are fed into SOTA \textbf{m}ulti-\textbf{m}odal \textbf{l}arge language \textbf{m}odels (MMLM($\cdot$), \emph{e.g.}, GPT~\cite{chatgpt}).
The process of UND can be represented as Eq.~\ref{eq:orc}.
\begin{equation}\small
\begin{aligned}
&\hspace{1.45cm}\underset{\Downarrow}{\underbrace{\left[\mathrm{OD}^1(\textbf{I}), \mathrm{OD}^2(\textbf{I}), \dots, \mathrm{OD}^k(\textbf{I})\right]}}\\[-0.2cm]
{\rm UNC_{\mathit{j}}}&\gets{\rm Man\left[MMLM\left(Pmt, \overline{O_\mathit{i}}, M_{\mathit{i}}\odot \mathcal{\textbf{I}}\right)\right]},\\
{\rm GT_{\mathit{i}}}&\gets{\rm Man\big[\underset{\Uparrow}{\underline{M_{\mathit{i}}}}\big]},\\[-0.2cm]
&\hspace{0.52cm}\overbrace{\rm VFM(\mathcal{\textbf{I}}, \underset{\Uparrow}{\underline{\rm BB_{i}}})}\\[-0.2cm]
&\hspace{-0.16cm}\overbrace{\rm \left[\mathrm{OD}^1(\textbf{I}), \mathrm{OD}^2(\textbf{I}), \dots, \mathrm{OD}^\emph{k}(\textbf{I})\right]}
\label{eq:orc}
\end{aligned}
\end{equation}
where $\mathcal{\textbf{I}}$, UNC$_{\mathit{j}}$, and GT denote each image in existing sets, \emph{j}-th user need command,
and the corresponding GT, respectively.
$\left[\mathrm{OD}^1(\cdot), \mathrm{OD}^2(\cdot), \dots, \mathrm{OD}^\emph{k}(\cdot)\right]$ denotes a set of existing object detectors---\emph{e.g.}, YOLOv8~\cite{yolov8}---are fine-tuned by different datasets from existing research in various fields (\emph{k}$\in$ (1, $+ \infty$)).
For a given image containing \emph{i}$\in$(1, $+ \infty$) objects (determined by the set of OD), each object has \emph{j} user need commands (determined by MMLMs, and manual correction for human expression-logic inconsistencies commands), where the \emph{j}-th commands correspond to the Ground Truth (GT$_i$) of the \emph{i}-th object.
MMLM($\cdot$), $\odot$, and VFM($\cdot$) denote multi-modality large language models, element-wise multiplication, and visual foundational models, respectively.
${\rm BB}_{i}$ and $\rm M_{\mathit{i}}$ denote object' bounding box and masks, respectively.
Pmt and Man[$\cdot$] denote extra prompt and correction in manual way.
By leverage Eq.~\ref{eq:orc}, UND can dig enough user need command and corresponding GT for existing samples (Fig.~\ref{fig:sets}(B)), and existing samples pairs from ($\mathcal{\textbf{I}}$, GT$^{b}$)$\rightarrow$($\mathcal{\textbf{I}}$, UNC$_{\mathit{j}}$, GT${\rm}_i$).
GT$^{b}$ denotes the GT of existing single sample pairs.
\subsection{Model and Loss: Fine-grained Usage}
\label{sec:Usersal}
\begin{figure}[!t]
	\centering
	\includegraphics[width=\linewidth]{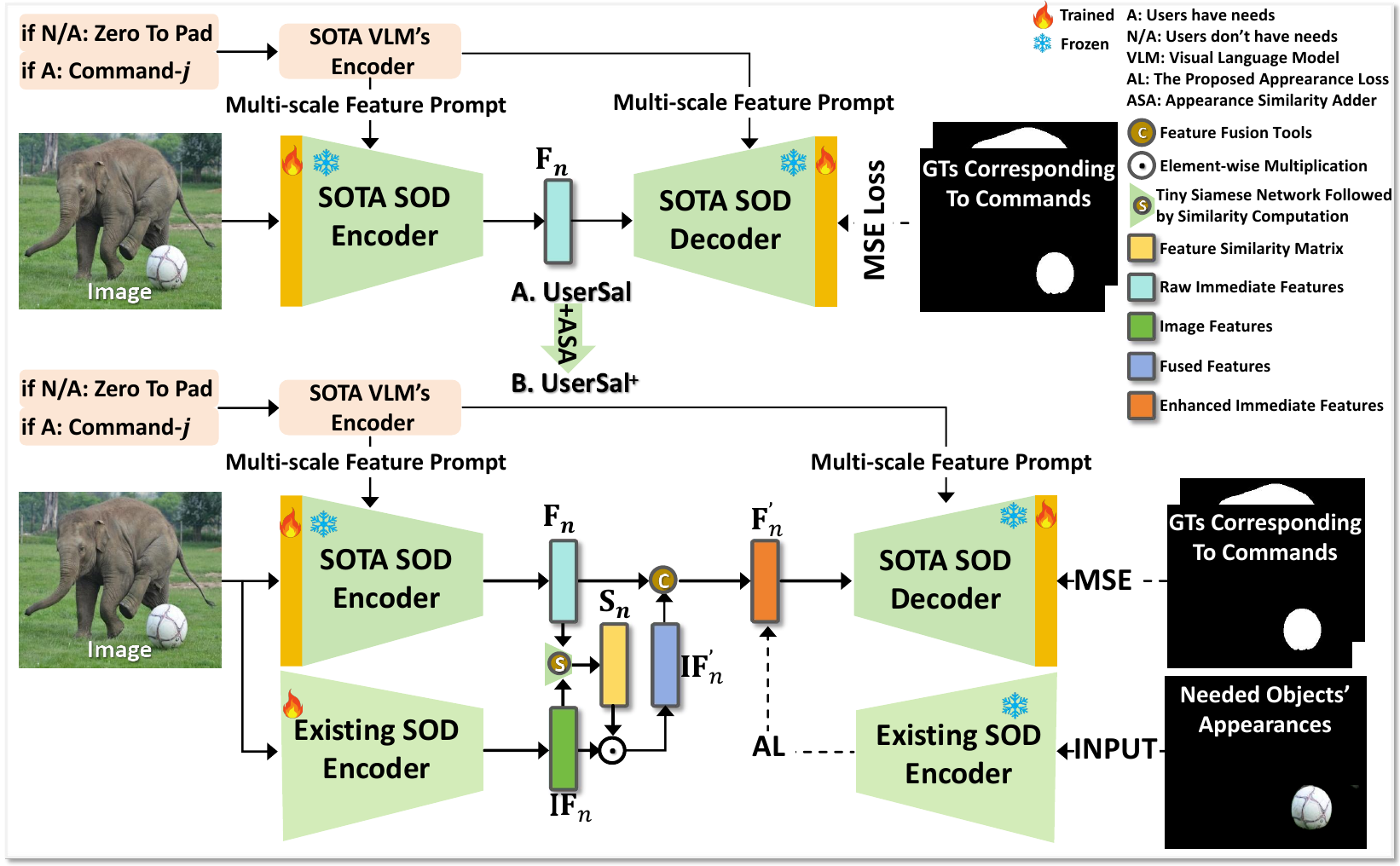}
	\caption{
		The proposed user need-driven model: Usersal (A) and Usersal$^+$ (B).}
	\label{fig:usersal}
\end{figure}
\label{sec:model}
In previous sections, we can obtain enough training and testing samples pairs.
However, \textbf{e}xisting \textbf{S}OD \textbf{m}odels (ESM) can't take multi-modal data (image \emph{v.s.} image and user need command text) as input.
Thus, we should address the second challenging problem---\textbf{how to train existing SOD model can leverage inputs (image and user need command text)?}
An intuitive solution is to leverage well-known \textbf{m}ulti-modal \textbf{p}rompt \textbf{l}earning (MPL) tool.
We propose a model: Usersal (Fig.~\ref{fig:usersal}(A)), which takes multi-scale text feature obtained by SOTA \textbf{v}ision \textbf{l}anguage models' \textbf{e}ncoders (VLE($\cdot$), \emph{e.g.}, BLIP~\cite{BLIP}) as \textbf{u}ser need \textbf{c}ommand \textbf{p}rompt (UCP$_j$) for frozen ESM append extra token.
The MPL of Usersal is represented as Eq.~\ref{eq:usersal}.
\begin{equation}\small
\begin{aligned}
&\hspace{0.9cm}\underset{\Downarrow}{\underbrace{\rm ESM^* +Tokens}}\\[-0.2cm]
&{\rm UserSal\gets \overline{ESM^{'}}\left(\mathcal{\textbf{I}}, \underline{UCP_\emph{j}}\right)\underset{MSE}{\Leftrightarrow} GT_{\mathit{i}},}\\[-0.2cm]
&\hspace{2.3cm}\underset{\overbrace{\rm VLE(URC_{\mathit{j}})}}{\Uparrow}
\label{eq:usersal}
\end{aligned}
\end{equation}
where UserSal, ESM$^{'}$, and ESM$^*$ denote our UserSal, existing frozen SOD method appended extra tokens, and frozen existing frozen SOD method, respectively.
$\mathcal{\textbf{I}}$, URC$_{\mathit{j}}$, and UCP$_j$ denote image, \emph{j}-th user need command, and multi-scale user need command prompt.
VLE($\cdot$) and MSE denote SOTA vision language models' encoders and the typical mean square error loss, respectively.
By Eq.~\ref{eq:usersal}, the Usersal can take image and user need command text as input, working well for user coarse-grained needs (\emph{e.g.}, I want find a people).

\begin{figure}[t]
	\centering
	\includegraphics[width=1\linewidth]{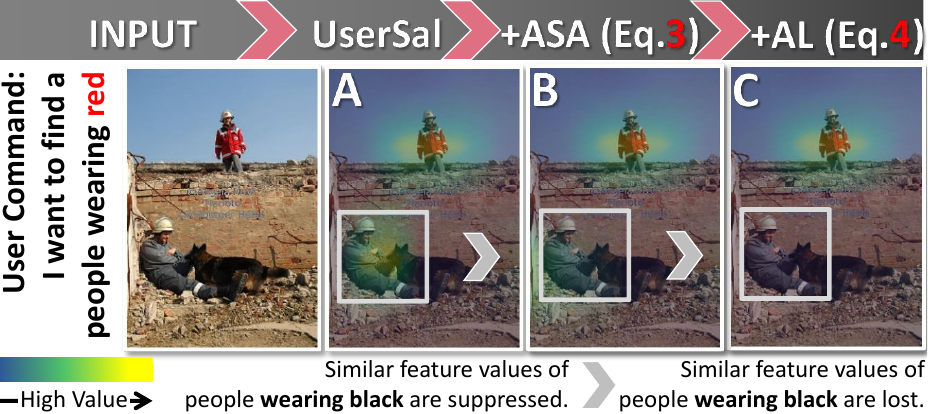}
	\caption{Visualization of similar feature.}
	\label{fig:rationale}
\end{figure}

But, as for users with fine-grained needs (\emph{e.g.}, I want to find a people wearing red.~\emph{v.s.}~I want find a people.), Usersal fails.
This is because existing MPL tools only train a model to represent multi-scale immediate feature $\textbf{\rm F}_\emph{n}$ with coarse-grained semantic similar information (Fig.~\ref{fig:rationale}(A)),
losing fine-grained appearance similar information.
Thus, we should deal with the third problem---\textbf{how to add fine-grained appearance similar information to the model trained by MPL}?
A component named \textbf{a}ppearance \textbf{s}imilarity \textbf{a}dder (ASA) is advised for Usersal, resulting in model--Usersal$^{+}$.

As shown in Fig.~\ref{fig:pipeline}(B), ASA consists of an existing \textbf{S}OD \textbf{m}ethod's \textbf{e}ncoder (SME($\cdot$)), a \textbf{s}imilarity \textbf{c}omputer (SC($\cdot$)), and an \textbf{a}ppearance \textbf{i}nformation \textbf{a}dder (AIA($\cdot$)).
Taking image $\mathcal{\textbf{I}}$ within input sample pairs, the SME encodes whole appearance information for it, outputting multi-scale image features \textbf{IF}$_\emph{n}$.
Then, to highlight user-needed targets' appearance information in \textbf{IF}$_\emph{n}$,
SC computes a similarity matrix between \textbf{IF}$_\emph{n}$ and \textbf{F}$_n$ by a \textbf{t}iny \textbf{S}iamese \textbf{n}etwork (TSN($\cdot$), \emph{e.g.}, single-layer self-attention SwinTranformer) followed \textbf{s}imilarity \textbf{c}omputation tool (\emph{e.g.}, cosine function), getting similarity matrix \textbf{S}$_\emph{n}$;
then, ASA element-wisely multiplies \textbf{S}$_\emph{n}$ with \textbf{IF}$_\emph{n}$, outputting features \textbf{IF}$^{'}_n$.
Finally, the AIA adds appearance similar information to \textbf{F}$_n$ by a \textbf{e}xisting feature \textbf{f}usion \textbf{t}ools (\emph{e.g.}, cross-attention) performed on \textbf{IF}$^{'}_n$ and \textbf{F}$_n$, outputting feature \textbf{F}$^{'}_n$.
For immediate features \textbf{F}$_n$ each scale (\textit{n}$\in$(1, N), generally, N=5), the process of ASA can be represent as Eq.~\ref{eq:ASD}.
\begin{equation}
\begin{aligned}
&\hspace{0.1cm}{\rm \mathbf{F}^{'}_\textit{n} \gets AIA(\mathbf{F}_\textit{n}, \underset{\Uparrow}{\underline{\mathbf{IF}{'}_\textit{n}})}},\\[-0.2cm]
&\hspace{2cm}\overbrace{\underset{\Uparrow}{\underline{\mathbf{S}_n}}\odot \mathbf{IF}_n}\\[-0.2cm]
&\hspace{0.7cm}\overbrace{\rm SC\left[TSN\left(\mathbf{F}_n, \underline{\mathbf{IF}_\textit{n}}\right)\right]}\\[-0.18cm]
&\hspace{2.5cm}\underset{\overbrace{\rm SME\left(\mathcal{\textbf{I}} \right)}}{\Uparrow}\\
\label{eq:ASD}
\end{aligned}
\end{equation}
where $\mathcal{\textbf{I}}$, \textbf{F}$_n$, \textbf{IF}$_n$, \textbf{S}$_n$, \textbf{IF}$^{'}_n$, and \textbf{F}$^{'}_n$
denote the image, Usersal (Eq.~\ref{eq:usersal})' immediate features, image features, feature similarity matrix, features getting by multiplying $\textbf{\rm S}_n$ by $\textbf{\rm IF}_n$,
and Usersal$^{+}$'s immediate features, respectively.
AIA($\cdot$), TSN($\cdot$), SC[$\cdot$], and SME($\cdot$) denote the appearance information adder, the tiny siamese network, the existing similarity computation tool, and the existing SOD method's encoder in ASA, respectively.
$\odot$ denotes element-wise multiplication.
Based on the ASA, we can add appearance similar information to Usersal's immediate features, which can suppress the dissimilar information about user fine-grained needs (Fig.~\ref{fig:rationale}(B)).
However, without a corresponding loss function, the effectiveness of ASA cannot be fully explored.
\begin{table*}[!t]
	\centering
	\begin{tabular}{c}
		\begin{minipage}{1\linewidth}
			\hspace{-8pt}
			\includegraphics[width=\linewidth]{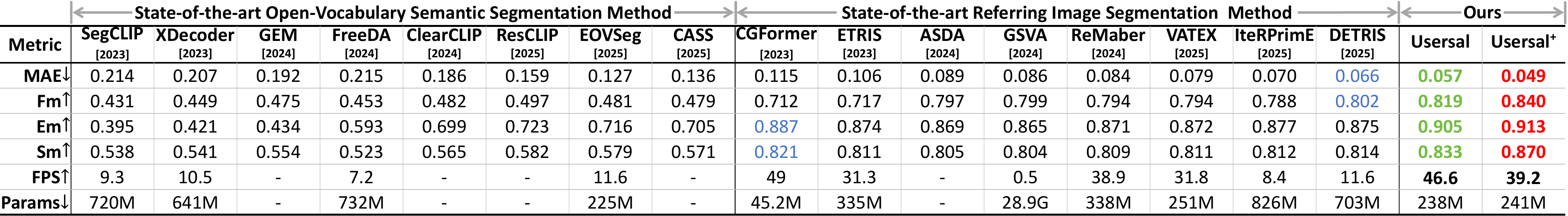}
		\end{minipage}
	\end{tabular}
	\caption{Quantitative comparisons between our method with other SOTA models on our proposed UserSOD set.
	The top three precision are highlighted with red, green and blue fonts.}
	\label{tab:aimcompare}
\end{table*}

Thus, to guide the ASA can to correct appearance similar information, we devise an appearance loss (AL).
AL first splites user-needed object appearance by element-wisely multiply GT$_i$ with image $\mathcal{\textbf{I}}$.
Then, a frozen encoder that is the same as existing SOD method's encoder (SME$^{*}$) and Kullback-Leibler Divergence loss (KL($\cdot$)) are leveraged to compute loss.
For each scale \textbf{F}$^{'}_n$, the AL can be represent as Eq.~\ref{eq:AL}.
\begin{equation}
\label{eq:AL}
\begin{aligned}
{\rm AL_\emph{n}\gets KL\left[SME^{*}(\textbf{I}\odot GT_\emph{i}), \mathbf{F}^{'}_\emph{n}\right]},
\end{aligned}
\end{equation}
where AL and KL[$\cdot$] denote our appearance loss and Kullback-Leibler Divergence loss, respectively.
$\odot$ and SME$^{*}(\cdot)$ denote element-wise multiplication and the frozen encoder that is the same as existing SOD method's encoder.
\textbf{F}$^{'}_n$, GT$_i$, $\mathcal{\textbf{I}}$ denote the feature outputted by the ASA, ground truth, and image, respectively.
With the AL, the effectiveness of the ASA can be fully explored, which enables immediate feature to only contain correct appearance similar information (Fig.~\ref{fig:rationale}(C)).
Based on AL, for $\mathit{j}$-th user need command (UNC$_{j}$) and $n$-th scale, our overall loss ($\mathcal{L}$) can be detailed as Eq.~\ref{eq:loss}.
\begin{equation}
\begin{aligned}
\label{eq:loss}
&{\rm \mathcal{L}\gets \left(\sum_{n=1}^{N}AL_\emph{n} + MSE\left[UserSal^+\left(\mathcal{\textbf{I}}, UNC_\emph{j}\right), GT_{\emph{i}}\right]\right)},
\end{aligned}
\end{equation}
where AL and MSE[$\cdot$] denote our appearance loss and the mean squared error loss, respectively.
$\mathcal{\textbf{I}}$ and GT$_i$ denote image and the GT corresponding to the \emph{j}-th user need command.

%% file: sec/4_Experiment.tex
\section{Experiment}
\subsection{Implementation Details}
We have trained our model by PyTorch with two NVIDIA GeForce 4090 GPU.
Our model is trained on UserSOD, and the input image is resized to 352$\times$352 resolution to accommodate the input needs of the model.
The learning rate is set to 1e-4 for the Adam optimizer and is later decayed by 10 at 60 epochs.
\subsection{Evaluation Metrics}
We utilize widely-used evaluation metrics for fair comparison, including \textbf{S}-measure (Sm)~\cite{Fan-ICCV17-Str}, \textbf{F}-measure (Fm)~\cite{Achanta-CVPR09-Fre}, \textbf{E}-measure (Em)~\cite{Fan-IJCAI18-Enh}, \textbf{m}ean \textbf{a}bsolute \textbf{e}rror (MAE), Parameters (Params), and running speed (FPS).
\subsection{Compare with SOTA Method}
\textbf{Evaluation on Our UserSOD Sets.}
We compare our method with other methods with image-text modality input, including 8 SOTA Open-Vocabulary Semantic Segmentation methods (OVSS, including SegCLIP~\cite{segclip}, XDecoder~\cite{xdecoder}, GEM~\cite{GEM}, FreeDA~\cite{FreeDA} ClearCLIP~\cite{ClearClip}, EOVSeg~\cite{EOVSeg}, and CASS~\cite{CASS}) and 8 Referring Image Segmentation methods (RIS, including CGFormer~\cite{CGFormer}, ETRIS~\cite{ETRIS}, ASDA~\cite{ASDA}, GSVA~\cite{GSVA}, ReMaber~\cite{ReMaber}, VATEX~\cite{VATEX}, IteRPrimE~\cite{IteRPrimE}, and DETRIS~\cite{DETRIS}) on the UserSOD set.
The comparison results in Table~\ref{tab:aimcompare} demonstrate that our method outperforms other SOTA models.
Compared with SOTA models, our method achieves on average 10\% improvement across four metrics.
This is because these models fail to leverage appearance similarity information from fine-grained user needs.
In contrast, our model can leverage this kind of information via the ASA (Eq.~\ref{eq:ASD}). 
\begin{table}[!b]
	\centering
	\begin{tabular}{c}
		\begin{minipage}{1\linewidth}
			\hspace{-8pt}
			\includegraphics[width=\textwidth]{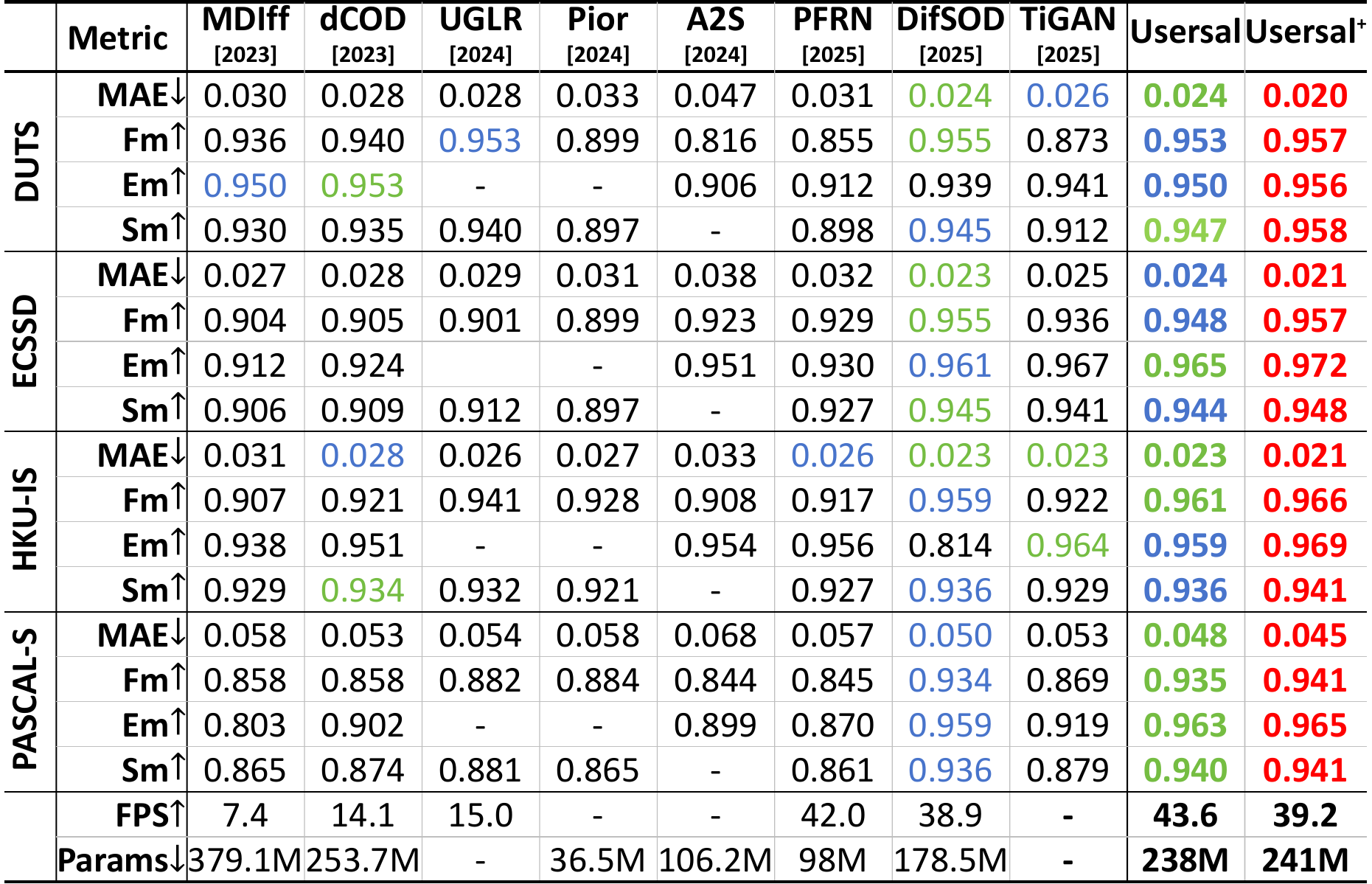}
		\end{minipage}
	\end{tabular}
	\caption{Quantitative comparisons between our method with SOTA SOD models on the conventional SOD sets.
	The top three precision are highlighted with red, green and blue fonts.}
	\label{tab:salcompare}
\end{table}

\begin{figure}[!t]
	\centering
	\includegraphics[width=1\linewidth]{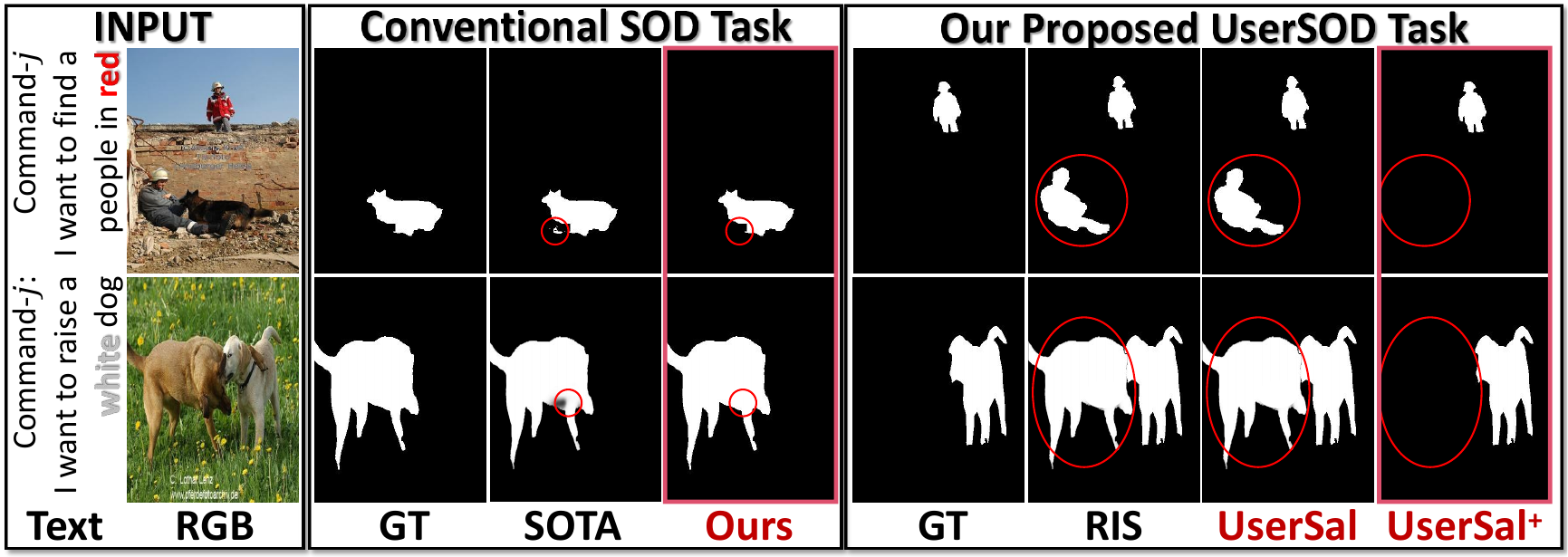}
	\caption{
		Visual comparisons between our SOTA \emph{v.s}. SOTA SOD and RIS methods. Our method not only achieves sharp contours for conventional SOD but also meets fine-grained user needs.}
	\label{fig:qual}
\end{figure}

\textbf{Evaluation on Existing SOD Sets.}
We compare our method with passive visual stimuli-based SOD methods (MDiff~\cite{Medsegdiff-v2}, dCOD~\cite{diffCOD}, UGLR~\cite{10489918}, Prior~\cite{Prior}, A2S~\cite{A2S}, PFRN~\cite{PFRN}, DifSOD~\cite{DiffSOD}, and TiGAN~\cite{TiGAN}) on SOD datasets, namely ECSSD~\cite{ECSSD}, HKU-IS~\cite{HKU-IS}, PASCAL-S~\cite{PASCAL-S}, and DUTS~\cite{DUTS} to evaluate our model's effectiveness for users without needs.
Due to the MPL's (Eq.~\ref{eq:usersal}) effectiveness and for fair comparison, in these comparisons, we only need to pad our model's text input with zeros, instead of re-training our model using conventional SOD sets.
Our method achieve new SOTA SOD performance (Table.~\ref{tab:salcompare}) for the main reasons:
1)~the UND (Sec.~\ref{sec:URD}) can supple more training sample, which are crucial for model's performance.
2)~our loss can guide model to learn fine-grained formation about salient objects, which are helpful for getting saliency maps with sharp contours
—these are strict problems for existing SOD methods
Additionally, we present qualitative evaluation results in Fig.~\ref{fig:qual}.

\begin{table}[!b]
	\centering
	\begin{tabular}{c}
		\begin{minipage}{1\linewidth}
			\hspace{-8pt}
			\includegraphics[width=\linewidth]{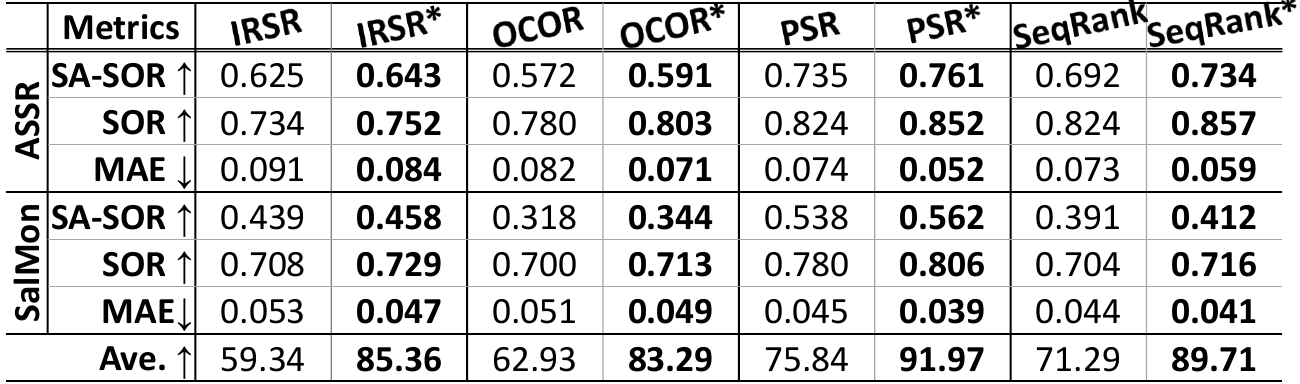}
		\end{minipage}
	\end{tabular}
	\caption{Our method's effectiveness for downstream SOR task.
$^{*}$ and Ave. denote baselines using our methods and average user satisfaction score, respectively.}
	\label{tab:Abl_SOR}
\end{table}
\textbf{Evaluation on Downstream Task.}
We choose a representative one among downstream tasks---salient object ranking (SOR)---to evaluate our model's effectiveness for downstream tasks.
Because most SOR methods highly rely on SOD methods, these methods take passive visual stimuli-based SOD methods as the core of their ``detecting salient object---erasing---detecting salient object'' iterative cycle ranking logic.
In these evaluations, we have taken SOTA SOR methods (IRSR~\cite{ISIR}, OCOR~\cite{OCOR}, PSR~\cite{PSR}, and SeqRank~\cite{SeqRank}) as baselines, and have replaced their passive visual stimuli-based SOD models with Usersal$^+$ to compare the performance of raw baselines and those equipped with Usersal$^+$.
As shown in Table~\ref{tab:Abl_SOR}, the performance comparison is divided into two aspect: A)~Comparison on SOR testing sets (ASSR~\cite{ASSR} and SalMon~\cite{SalMon}), B)~Comparison of User Satisfaction between baselines and those equipped with Usersal$^+$.
In general, performance comparisons prove that our UserSOD task and corresponding model can advance downstream tasks.

A)~
For successful comparison and to avoid data leakage from the SOR testing dataset, we input the appearance of split objects that rank the highest in GT into MMLMs that are different from those in UND (Unlike our method, it prevents model access to test set data and avoids data leakage) (Eq.~\ref{eq:orc}), obtaining corresponding user need commands. 
Then, the whole images and corresponding user need commands are fed into baselines equipped with Usersal$^+$ to get TOP-5  (this setting is consistent with baselines) ranking results.
As shown in Table~\ref{tab:Abl_SOR}, compared to raw baselines, those equipped with Usersal$^+$ have better performance 
across both SOR testing datasets, \textit{i.e.}, higher SA-SOR and SOR scroes, and lower MAE.

B)~
The user studies for evaluating our model's effectiveness in SOR are recruit 100 participants (different from the evaluation for the URD, different jobss, gender, personalities, etc.) with their objectives, \emph{i.e.}, shopping, industry, and navigation.
All Images contained in ASSR and SalMon corresponding with these user's need commands and those images are feed to the baseline equipped Usersal$^+$ and raw baselines, respectively, and then users score (0-100) to both methods' ranking results.
As can been seen in Table~\ref{tab:Abl_SOR}, compared to raw baselines, those equipped Usersal$^+$ can satisfy users more effectively, \emph{i.e.}, achieving more higher average user satisfaction score.
The reason of these advancements is that compared to conventional passive visual stimuli-based SOD models, our model makes these SOR methods obtain the ability to analysis fine-grained relationship between user's viewing behavior and needs, which is necessary for SOR methods.

\textbf{Generalization Evaluation on RIS Sets and User Study.}
The RIS can be regraded as a special case of UerSOD task, \emph{i.e.}, inputting scenes that contain user-required salient objects;
thus, the sets of RIS (RefCoCo+~\cite{RefCoCo} and RefCoCo++~\cite{RefCoCo}) can used to evaluated our method's generalization on other tasks.
The results are reported in Table~\ref{tab:generic}-\ding{172}.
Compared to existing SOTA methods, ASA and AL can make our model can leverage fine-grained information (\emph{i.e.}, appearances) to segment objects.
Thus, our method outperform SOTA RIS method.
\begin{table}[t]
	\centering
	\begin{tabular}{c}
		\begin{minipage}{1\linewidth}
			\hspace{-8pt}
			\includegraphics[width=\linewidth]{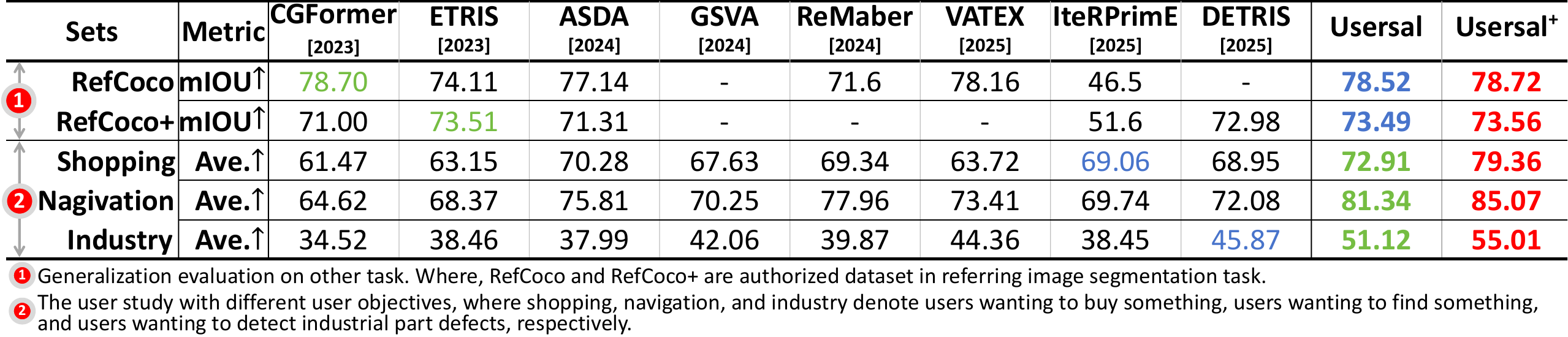}
		\end{minipage}
	\end{tabular}
	\caption{Generalization evaluation on RIS Sets and user study.
	The top three precision are highlighted with red, green and blue fonts.
Where, Ave. and mIOU denotes average user satisfaction score and widely-used metric on RIS sets, respectively.}
    \label{tab:generic}
\end{table}

The user studies recruited 100 participants (different from those in the evaluations for the UND and the effectiveness of SOR, with different jobs, genders, personalities, etc.) with their objectives, \emph{i.e.}, shopping, industry, and navigation.
3,000 inputting image (including 1,000 shopping, 1,000 navigation, and 1,000 industry) are collected from Youtube.
As shown in Table~\ref{tab:generic}-\ding{173}, compared to existing methods, our method obtains the highest average user satisfaction score.
However, compared to shopping and navigation, users with industry-specific objectives have significantly lower satisfaction.
This explores the \textbf{limitations} of our approach.
Due to the lack of industry-related samples, we are unable to fully explore industrial user needs, resulting in insufficient industry adaptability of our method.
However, we can leverage the UND to mine samples from industrial task sets (\emph{e.g.}, defect detection, anomaly detection, etc.) to mitigate this limitation.

\subsection{Component Evaluation}
\textbf{The effectiveness of UND.}
We need to verify whether the needs dug by UND:
1)~Align with genuine user needs; 2)~commands being similar to needed objects.
To this end, we randomly selected 100 sample pairs from the UserSOD set, each containing an image, user need commands, and the corresponding GT.
Based on these samples, we have conducted a questionnaire survey with 100 volunteers to rate (0-100), covering the following aspects: 1)~Do you believe you would have the same need in the given scenario?
2)~Does the need command text follow natural human language expression? 3)~When the \emph{j}-th need is present, do you primarily focus on the segmented object in the image?
4)~Do you think the need command text aligns with the segmented object?
The results (average rate: 100, 100, 100, and 100 and variance: 0, 0, 0 and 0 for each question) indicate that the user need commands dig by the UTD align well with user needs.
\begin{table}[!t]
	\centering
	\begin{tabular}{c}
		\begin{minipage}{1\linewidth}
			\hspace{-8pt}
			\includegraphics[width=\linewidth]{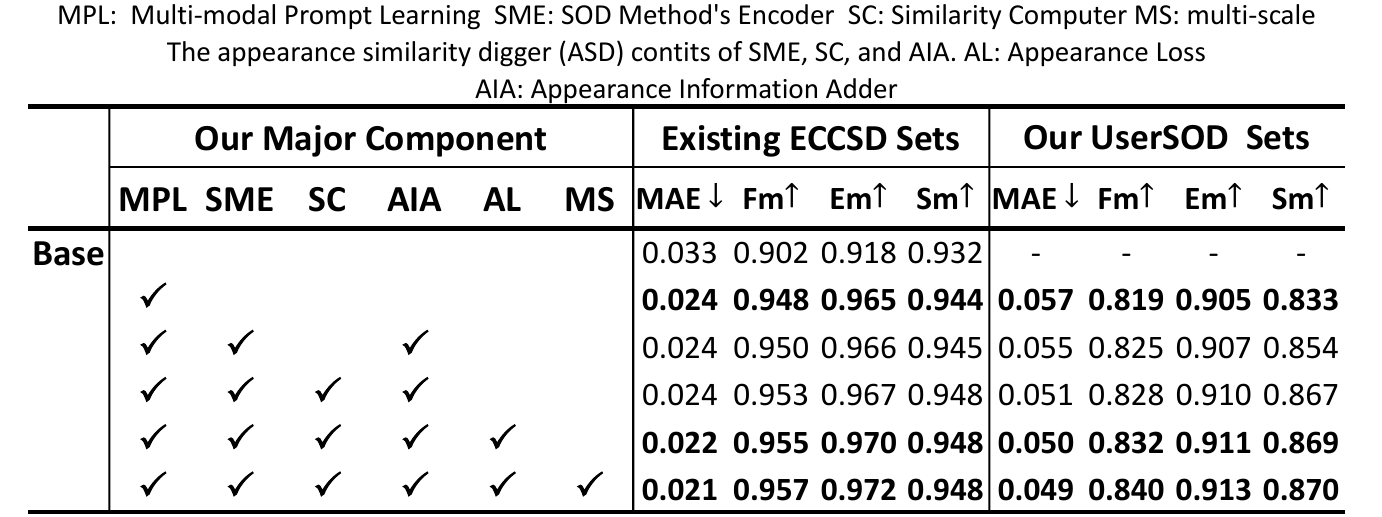}
		\end{minipage}
	\end{tabular}
	\caption{Component evaluation on conventional SOD and UserSOD sets.}
	\label{tab:Component}
\end{table}

Furthermore, to validate whether the commands are similar to the needed objects, we input the user need commands into existing vision-language foundation models (VLMs, \emph{e.g.}, CLIP) and generative models (\emph{e.g.}, Stable Diffusion; note: UND does not incorporate any VLMs, making it suitable for independent evaluation).
We have computed the cosine similarity between the features of the generated objects and the needed objects extracted from the images by a well-trained ViT.
The average similarity scores and variances are 0.995 and 0.583, respectively, proving the user need commands obtained by UND correspond to the needed objects.

\textbf{The effectiveness of the MPL.}
We take VST~\cite{Liu-arxiv21-VST} as the baseline (Table~\ref{tab:Component}(Base)).
By taking text feature obtained from existing MMLM as prompt,
the MPL can make the baseline to perform the UerSOD task (Table~\ref{tab:Component}(\ding{172})).
Meanwhile, due to our UND can obtain more training samples, it boosts baseline's performances on the SOD task.

\textbf{The effectiveness of the ASA.}
The results in Table~\ref{tab:Component}(\ding{173}-\ding{174}) show that
our appearance information adder can supply user-needed target appearance similar information for our method, which is proven by performance across both sets.

\textbf{The Effectiveness of the AL.}
The propose of that our proposed loss is to guide ASA to add correct user-needed targets' appearance similarity information.
Table~\ref{tab:Component}-\ding{175} shows that our proposed loss significantly improves the model's overall performance, which  proves our idea's effectiveness.
Finally, we verify applying our each component for multi-scale performing (Table~\ref{tab:Component}-\ding{176}), the overall performance of our method improve again, which proves that performing each component at multi-scale is effective.
\begin{table}[t]
	\centering
	\begin{tabular}{c}
		\begin{minipage}{1\linewidth}
			\hspace{-8pt}
			\includegraphics[width=\linewidth]{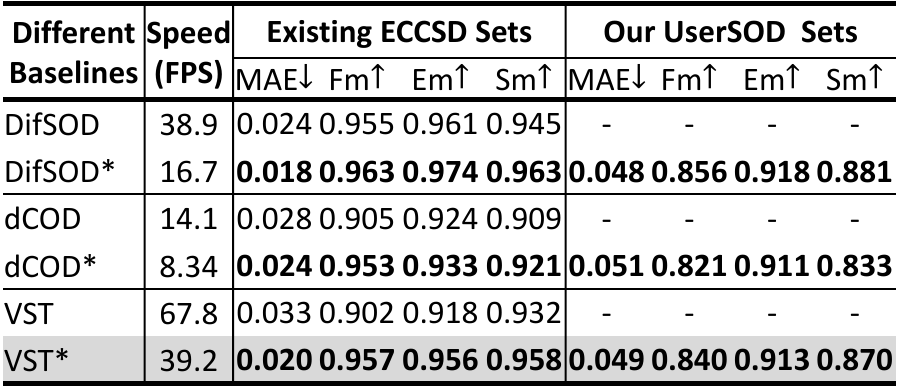}
		\end{minipage}
	\end{tabular}
	\caption{Generalization evaluation for different baselines on conventional SOD and UserSOD sets.
$^{*}$ denote baselines using our methods.}
	\label{tab:ablbaseline}
\end{table}
\subsection{Ablation Study}
\textbf{Different Baseline.}
We evaluate our method's generalization across different baselines (Table~\ref{tab:ablbaseline}).
For fair comparison, we taking VST, DifSOD, and dCOD as different baselines, using our training set and method to train them.
\begin{table}[!b]
	\centering
	\begin{tabular}{c}
		\begin{minipage}{1\linewidth}
			\hspace{-8pt}
			\includegraphics[width=\linewidth]{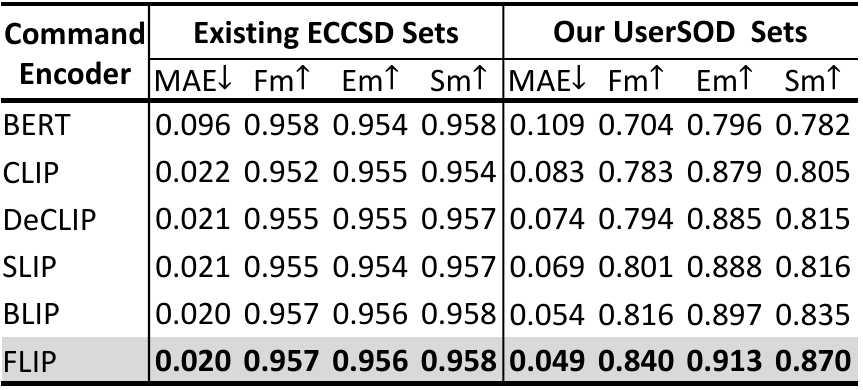}
		\end{minipage}
	\end{tabular}
	\caption{Different command encoder evaluation on conventional SOD and UserSOD sets.}
	\label{tab:abltextencoder}
\end{table}
Compared to baselines, our method improve their performance across both conventional SOD sets.
Although DifSOD* (* denotes the performance of baseline after using our method) achieve best performance, it is too slow (16.7FPS).
Thus, we take VST*'s performance as our final reported performance.

\textbf{Different User Need Command Encoder.}
We explore how different SOTA VLMs' encoders (\textit{i.e.}, User Need Command Encoder) affect our model's performance (Table~\ref{tab:abltextencoder}).
In these comparisons, the SOTA VLMs' encoders including BERT~\cite{BERT}, CLIP, DeCLIP~\cite{DeCLIP}, SCLIP~\cite{SCLIP}, BLIP, and FLIP are taken as our user need command encoders.
Among them, BERT-based variant achieves the lowest performance.
This is because its encoder failed to output features in the same feature space as the image encoder.
In contrast, other encoders leverage contrastive learning and similar methods to align text and image features domains, thereby achieving better performance.
Additionally, these approach also resolve the issue of multi-modal feature domain gaps in our model.
On conventional SOD sets, the user need commands are padded with zeros (rendering these encoders ineffective), and their different variants exhibited similar performance in SOD task, the FLIP-based variant achieved the best results in UserSOD task.
Therefore, we adopt it as the basis for our performance reporting.

\textbf{Different Tiny Siamese Network.}
We explore how different tiny Siamese networks including single-layer fully connect, convolution, self-attention in ViT~\cite{VIT}, and self-attention in SwinTransformer~\cite{Swin}) affect our model's performance (Table~\ref{tab:abltinysiamese}).
Among them, the self-attention in SwinTransformer due to shifted window mechanism variant get the best performance.
\begin{table}[!t]
	\centering
	\begin{tabular}{c}
		\begin{minipage}{1\linewidth}
			\hspace{-8pt}
			\includegraphics[width=\linewidth]{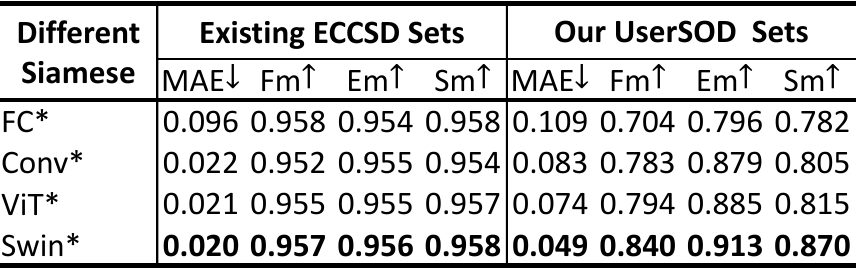}
		\end{minipage}
	\end{tabular}
	\caption{Tiny siamese network evaluation on conventional SOD and UserSOD sets.
$^{*}$ denotes single-layers.}
	\label{tab:abltinysiamese}
\end{table}

%% file: sec/5_Concluison.tex
\section{Conclusion}
In this paper, we introduce a new task, named UserSOD, which emphasizes the importance of use need for detecting salient object that existing SOD tasks  overlook.
To address the issue of sample shortage, we propose a user need digger, which can dig user need for existing sample, cutting down on labor and money.
Besides, we propose a model and corresponding loss that can meet user fine-grained need.
Our contributions can drive advancements in the field.